\documentclass[11pt,a4paper]{article}
\usepackage{times,latexsym}
\usepackage{url}
\usepackage[T1]{fontenc}

\usepackage[acceptedWithA]{tacl2018v2}
\usepackage{graphicx}
\usepackage{booktabs}
\usepackage{multirow}
\usepackage{multirow}
\usepackage{multirow}
\usepackage{multirow}
\usepackage{multirow}
\usepackage{multirow}
\usepackage{multirow}
\usepackage{multirow}
\usepackage{multirow}
\usepackage{multirow}
\usepackage{multirow}
\usepackage{multirow}
\usepackage{multicol}
\usepackage{makecell}
\usepackage{enumitem}

\usepackage{xspace,mfirstuc,tabulary}

\newcommand{\mb}{\mathbf}

\newif\iftaclinstructions
\taclinstructionsfalse 
\iftaclinstructions
\renewcommand{\confidential}{}
\renewcommand{\anonsubtext}{(No author info supplied here, for consistency with
TACL-submission anonymization requirements)}
\newcommand{\instr}
\fi

\iftaclpubformat 

\newcommand\rct{\textsc{RCT}\xspace}

\newcommand\chemprot{\textsc{ChemProt}\xspace}
\newcommand\arccite{\textsc{ACL-ARC}\xspace}
\newcommand\sciie{\textsc{SciERC}\xspace}
\newcommand\hp{\textsc{HyperPartisan}\xspace}

\newcommand\ag{\textsc{AGNews}\xspace}
\newcommand\helpful{\textsc{Helpfulness}\xspace}
\newcommand\imdb{\textsc{IMDB}\xspace}

\newcommand\news{\textsc{News}\xspace}
\newcommand\med{\textsc{BioMed}\xspace}
\newcommand\cs{\textsc{CS}\xspace}

\newcommand\reviews{\textsc{Reviews}\xspace}
\else

\fi

\title{Self-supervised Regularization for Text Classification}

\author{Meng Zhou${^*}$ \\ Shanghai Jiao Tong University \\  {\tt\small zhoumeng9904@sjtu.edu.cn} 
        \And  Zechen Li \Thanks{Equal Contribution} \\ Northeastern University \\ {\tt\small li.zec@northeastern.edu} \And
        Pengtao Xie\Thanks{Corresponding Author} \\ UC San Diego \\ {\tt\small p1xie@eng.ucsd.edu}}
\date{}

\begin{document}
\maketitle
\begin{abstract}
Text classification is a widely studied problem and has broad applications. In many real-world problems, the number of texts for training classification models is limited, which renders these models prone to overfitting. To address this problem, we propose SSL-Reg, a data-dependent regularization approach based on self-supervised learning (SSL). SSL~\cite{devlin2018bert} is an unsupervised learning approach which defines auxiliary tasks on input data without using any human-provided labels and learns data representations by solving these auxiliary tasks. In SSL-Reg,  a supervised classification task and an unsupervised SSL task are performed simultaneously. The SSL task is unsupervised, which is defined purely on input texts without using any human-provided labels. Training  a model using an SSL task can prevent the model from being overfitted to a limited number of class labels in the classification task. 
 Experiments on 17 text classification datasets demonstrate the effectiveness of our proposed method. Code is available at \url{https://github.com/UCSD-AI4H/SSReg}
\end{abstract}

\section{Introduction}
Text classification~\cite{Korde2012TEXTCA, Lai2015RecurrentCN, Wang2017CombiningKW, Howard2018UniversalLM} is a widely studied problem in natural language processing and finds broad applications. For example, give  clinical notes of a patient, judge whether this patient has heart diseases. Given a scientific paper, judge whether it is about NLP. 
In many real-world text classification problems,  texts available for training are oftentimes limited. For instance, it is difficult to obtain a lot of clinical notes from hospitals due to  concern of patient privacy. It is well known that when  training data is limited, models tend to overfit to  training data and perform less well on test data.   

To address  overfitting problems in text classification, we propose  a data-dependent regularizer called SSL-Reg based on self-supervised learning (SSL)~\cite{devlin2018bert,he2019moco,chen2020simple} and use it to regularize the training of text classification models, where a supervised classification task and an  unsupervised SSL task are performed simultaneously. Self-supervised learning (SSL)~\cite{devlin2018bert,he2019moco,chen2020simple} is an unsupervised learning approach which defines auxiliary tasks on input data without using any human-provided labels and learns data representations by solving these auxiliary tasks. For example,  BERT~\cite{devlin2018bert} is a typical SSL approach where an  auxiliary task is defined to predict masked tokens and a  text encoder is learned by solving this task. 
In existing SSL approaches for NLP, an SSL task and a target task are performed sequentially. 
A text encoder is first trained by solving an SSL task defined on a large collection of unlabeled texts. Then this encoder is used to initialize an encoder in a target task. The encoder is finetuned by solving the target task. A potential drawback of performing SSL task and  target task sequentially is that  text encoder learned in  SSL task may be overridden after being finetuned in target task. If  training data in the target task is small, the finetuned encoder has a high risk of being overfitted to training data.

To address this problem, in SSL-Reg we perform  SSL task and  target task (which is classification) simultaneously. In SSL-Reg, an SSL loss serves as a regularization term and is optimized jointly with a classification loss.    SSL-Reg enforces a text encoder to jointly solve two tasks: an unsupervised SSL task and a supervised text classification task. Due to the presence of the SSL task, models are less likely to be biased to the classification task defined on  small-sized training data. 
We perform experiments on 17 datasets, where experimental  results demonstrate the effectiveness of SSL-Reg in alleviating overfitting and improving generalization performance.

The major contributions of this paper are:
\begin{itemize}[leftmargin=*]
    \item We propose SSL-Reg, which is a data-dependent regularizer based on SSL, to reduce the risk that a  text encoder is biased to a data-deficient classification task on  small-sized training data.
    \item Experiments on 17 datasets demonstrate the effectiveness of our approaches. 
\end{itemize}

The rest of this paper is organized as follows. Section 2 reviews related works. Section 3 and 4 present  methods and experiments. Section 5 concludes the paper and discusses future work.

\section{Related Works}

\subsection{Self-supervised Learning for NLP}
Self-supervised learning (SSL) aims to learn meaningful representations of input data without using human annotations. It creates auxiliary tasks solely using  input data and forces deep networks to learn highly-effective latent features by solving these auxiliary tasks. In NLP, various auxiliary tasks have been proposed for SSL, such as next token prediction in GPT~\cite{radford2018improving}, masked token prediction in BERT~\cite{devlin2018bert}, text denoising in BART~\cite{lewis2019bart}, contrastive learning~\cite{fang2020cert}, and so on. These models have achieved substantial success in learning  language representations. 
The GPT model~\cite{radford2018improving} is a language model (LM) based on Transformer~\cite{vaswani2017attention}. Different from Transformer which defines a conditional probability on an output sequence given an input sequence, GPT defines a marginal probability on a single sequence. In GPT,  conditional probability of the next token given a historical sequence is defined using a  Transformer decoder. Weight parameters are learned by maximizing  likelihood on token  sequences.  
BERT~\cite{devlin2018bert} aims to  learn a Transformer encoder for representing texts. BERT’s model architecture is a multi-layer bidirectional Transformer encoder. In BERT,  Transformer uses bidirectional self-attention. To train the encoder, BERT masks some percentage of  input
tokens at random, and then predicts those masked tokens by feeding  hidden vectors (produced by the encoder) corresponding to  masked tokens into an output softmax over word vocabulary.
BERT-GPT~\cite{wu2019importance} is a model used for sequence-to-sequence modeling where a pretrained BERT is used to encode input text and GPT is used to generate output texts.
In BERT-GPT,  pretraining of  BERT encoder and  GPT decoder is conducted separately, which may lead to inferior performance.
Auto-Regressive
Transformers (BART)~\cite{lewis2019bart} has a similar architecture as BERT-GPT, but trains  BERT encoder and GPT decoder jointly. To pretrain  BART weights,  input texts are corrupted randomly, such as token masking, token deletion, text infilling, etc., then a  network is
learned to reconstruct  original texts.  ALBERT~\cite{lan2019albert} uses parameter-reduction methods to reduce  memory consumption and increase  training speed of BERT. It also introduces a self-supervised loss which models inter-sentence coherence. RoBERTa~\cite{liu2019roberta} is a replication study of BERT pretraining. It shows that  BERT's performance can be greatly improved by carefully tuning  training processes, such as (1) training  models longer, with larger  batches, over more data; (2) removing the next sentence prediction objective; (3) training on longer sequences, etc. XLNet~\cite{yang2019xlnet} learns bidirectional contexts by maximizing  expected likelihood over all permutations of  factorization order and uses a generalized autoregressive pretraining mechanism to overcome the pretrain-finetune discrepancy of BERT. T5~\cite{raffel2019exploring} compared pretraining objectives, architectures, unlabeled datasets, transfer approaches on a wide range of language understanding tasks and proposed a unified framework that casts these tasks as a text-to-text task. 
ERNIE 2.0~\cite{sun2019ernie} proposed a continual pretraining framework which builds and learns incrementally pretraining tasks through constant multi-task learning, to capture  lexical, syntactic and semantic information from training corpora. \citet{dontstoppretraining2020} proposed task adaptive pretraining (TAPT) and domain adaptive pretraining (DAPT). Given a RoBERTa model pretrained on large-scale corpora, TAPT continues to pretrain RoBERTa on training dataset of target task. DAPT continues to pretrain RoBERTa on datasets that have small domain differences with data in target tasks. The difference between our proposed SSL-Reg method with TAPT and DAPT is that SSL-Reg uses a self-supervised task (e.g., mask token prediction) to regularize the finetuning of RoBERTa where text classification task and self-supervised task are performed jointly. In contrast,  TAPT and DAPT use self-supervised task for pretraining, where text classification task and self-supervised task are performed sequentially. The connection between our method and TAPT is that they both leverage texts in target tasks to perform self-supervised learning, in addition to SSL on large-scale external corpora. Different from SSL-Reg and TAPT, DAPT uses domain-similar texts rather than target texts for additional SSL.

\subsection{Self-supervised Learning in General}
 
Self-supervised learning has been widely applied to other application domains, such as image classification~\cite{he2019moco,chen2020simple}, graph classification~\cite{zeng2021contrastive}, visual question answering~\cite{he2020pathological}, etc, where  various strategies have been proposed to construct auxiliary tasks, based on temporal correspondence~\cite{li2019joint,wang2019learning}, cross-modal consistency~\cite{wang2019reinforced},  rotation prediction~\cite{gidaris2018unsupervised, sun19ttt}, image inpainting~\cite{pathak2016context}, automatic colorization~\cite{zhang2016colorful}, context prediction~\cite{nathan2018improvements}, etc.
Some recent works studied self-supervised representation learning based on instance discrimination~\cite{wu2018unsupervised} with contrastive learning. \citet{oord2018representation}  proposed contrastive predictive coding (CPC), which  predicts the future in latent space by using powerful autoregressive models, to extract useful representations from high-dimensional data.  
\citet{bachman2019learning}  proposed a self-supervised representation learning approach based on maximizing mutual information between features extracted from multiple views of a shared context. 
MoCo~\cite{he2019moco}  and SimCLR~\cite{chen2020simple} learned image encoders by predicting whether two augmented images were created from the same original image. 
\citet{srinivas2020curl} proposed to learn  contrastive unsupervised representations for reinforcement learning. \citet{khosla2020supervised} investigated supervised contrastive learning, where clusters of points belonging to the same class were pulled together in embedding space, while clusters of samples from different classes were pushed apart. \citet{klein2020contrastive} proposed a contrastive self-supervised learning approach for commonsense reasoning. \citet{he2020sample,yang2020transfer} proposed an Self-Trans approach which applied contrastive self-supervised learning on top of networks pretrained by transfer learning.

Compared with supervised learning which requires each data example to be labeled by humans or semi-supervised learning which requires part of data examples to be labeled, self-supervised learning is similar to unsupervised learning because it does not need human-provided labels. The key difference between self-supervised learning (SSL) and unsupervised learning is that SSL focuses on learning data representations by solving auxiliary tasks defined on un-labeled data while unsupervised learning is more general and aims to discover latent structures from data, such as clustering, dimension reduction, manifold embedding~\cite{roweis2000nonlinear}, etc.

\subsection{Text Classification}

Text classification~\cite{minaee2020deep} is one of the key tasks  in natural language processing and has a wide range of applications, such as sentiment analysis, spam detection, tag suggestion, etc. A number of approaches have been proposed for text classification. Many of them are based on RNNs. 
\citet{liu2016recurrent} use multi-task learning to train RNNs, utilizing the correlation between tasks to improve text  classification performance. \citet{tai-etal-2015-improved} generalize sequential LSTM to tree-structured LSTM to capture the syntax of sentences for achieving better classification performance.  Compared with RNN-based models, CNN-based models are good at capturing local and position-invariant patterns. \citet{kalchbrenner-etal-2014-convolutional} proposed dynamic CNN (DCNN), which uses dynamic k-max-pooling to explicitly capture short-term and long-range relations of words and phrases. \citet{Zhang2015CharacterlevelCN} proposed a character-level CNN model for text classification, which can deal with out-of-vocabulary words.  Hybrid methods  combine RNN and CNN to explore the advantages of both. \citet{Zhou2015ACN} proposed a convolutional LSTM  network, which uses a CNN to extract  phrase-level  representations, then feeds them to an LSTM network to represent the whole sentence.

\section{Methods}
To alleviate overfitting in text classification, we propose SSL-Reg, which is a regularization approach based on self-supervised learning (SSL), where an unsupervised SSL task and a supervised text classification task are performed jointly. 

\begin{figure}[t]
    \centering
    \includegraphics[width=0.3\textwidth]{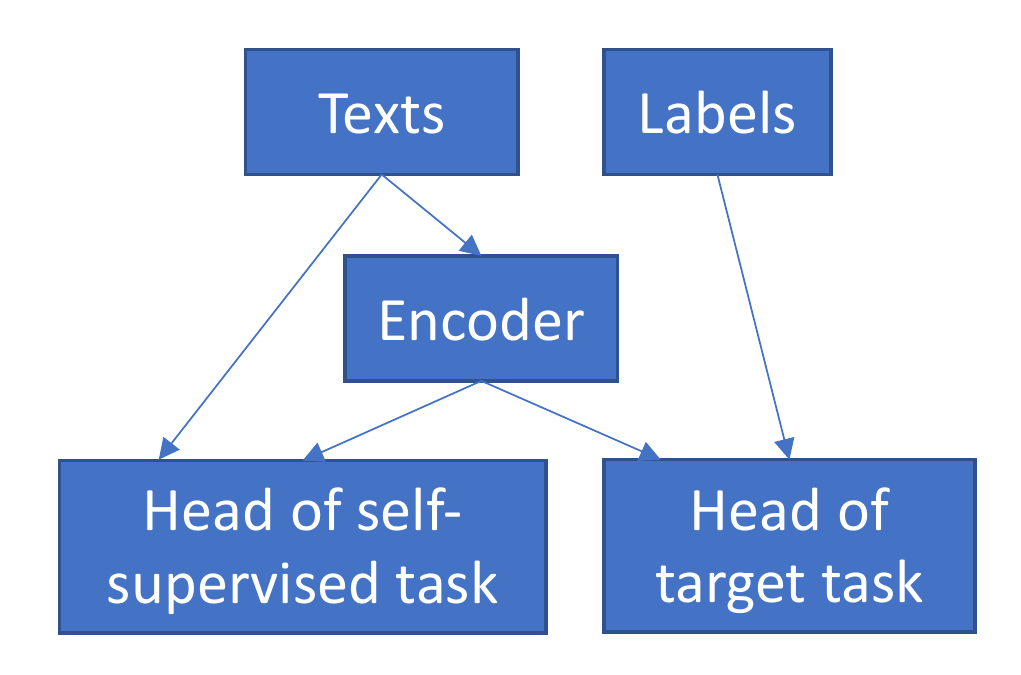}
    \caption{Illustration of SSL-Reg. Input texts are fed into a  text encoder.  Encodings of these texts and their corresponding labels are fed into the head of a target task (e.g.,  classification) which yields a classification loss. In a self-supervised task,  inputs are encodings of texts and outputs are constructed on original texts (e.g., masked tokens). The classification task and SSL task share the same text encoder and losses of these two tasks are optimized jointly to learn the text encoder.   
    }
    \label{fig:reg}
\end{figure}

\subsection{SSL-based Regularization}
SSL-Reg uses a self-supervised learning  task to regularize a  text classification model. Figure~\ref{fig:reg} presents an illustration of SSL-Reg.  Given  training texts, we encode them using a text encoder. Then on top of  text encodings, two tasks are defined. One is a classification task, which takes the encoding of a text as input and predicts the class label of this text. Prediction is conducted using a classification head. The other task is SSL.  The loss of the SSL task serves as a data-dependent regularizer to alleviate overfitting. The SSL task has a predictive head. These two tasks share the same text encoder. Formally, SSL-Reg solves the following optimization problem:
\begin{equation}
\label{eq:general}
\begin{array}{l}
    \mathcal{L}^{(c)}(D,L;\mb{W}^{(e)}, \mb{W}^{(c)})+ \lambda\mathcal{L}^{(p)}(D,\mb{W}^{(e)},\mb{W}^{(p)} )
    \end{array}
\end{equation}
where $D$ represents  training texts and $L$ represents their labels. $\mb{W}^{(e)}$, $\mb{W}^{(c)}$, and $\mb{W}^{(p)}$ denote text encoder, classification head in  classification task, and prediction head in  SSL task respectively. $\mathcal{L}^{(c)}$ denotes  classification loss and $\mathcal{L}^{(p)}$ denotes  SSL loss. $\lambda$ is a tradeoff parameter.

At the core of SSL-Reg is using SSL to  learn a text encoder that is robust to overfitting.  Our methods can be used to learn any text encoder. In this work, we perform the study using a Transformer encoder, while noting that other text encoders are also applicable.

\begin{table*}[t]
\centering
\begin{tabular}{lllrrrr}
\toprule
\textbf{Domain}
& \textbf{Dataset}            
& \textbf{Label Type} 
& \textbf{Train} 
& \textbf{Dev} 
& \textbf{Test} 
& \textbf{Classes}      \\
\midrule
\multirow{2}{*}{\med} 
& \chemprot 
& relation classification
& 4169  
& 2427  
& 3469 
&       13    \\
& \rct            
& abstract sent. roles 
& 180040    
& 30212 
& 30135 
& 5 \\
\midrule[0.03em]
\multirow{2}{*}{\cs}
& \arccite  
& citation intent 
& 1688    
&  114 
& 139 
& 6                     \\
& \sciie     
&  relation classification     
& 3219 
& 455  
& 974        
& 7    \\
\midrule[0.03em]
\multirow{2}{*}{\news}
& \hp 
& partisanship 
& 515 
& 65
& 65 
& 2             \\
& \ag       
& topic  
& 115000    
& 5000   
& 7600  
& 4      \\
\midrule[0.03em]
\multirow{2}{*}{\reviews} 
& \helpful
& review helpfulness 
& 115251 
& 5000 
& 25000  
& 2    \\
& \imdb     
& review sentiment  
& 20000 
& 5000 
& 25000 
& 2       \\ 
\bottomrule
\end{tabular}
\caption{ Statistics of datasets used in \citep{dontstoppretraining2020}. 
Sources: \chemprot \citep{Kringelum2016ChemProt30AG}, \rct \citep{Dernoncourt2017PubMed2R}, \arccite \cite{Jurgens2018MeasuringTE}, \sciie \citep{Luan2018MultiTaskIO}, \hp \citep{stein:2019h}, \ag \citep{Zhang2015CharacterlevelCN}, \helpful \citep{mcauley2015image}, \imdb \citep{Maas2011LearningWV}. This table is taken from \citep{dontstoppretraining2020}.
}
\label{tab:data_tasks}
\end{table*}

\begin{table*}[h]
    \centering
    \begin{tabular}{l|c|c|c|c|c|c|c|c|c|c}
    \toprule
       & CoLA & RTE & QNLI & STS-B  & MRPC & WNLI & SST-2 & \makecell{MNLI\\(m/mm)} & QQP & AX\\
        \hline
Train & 8551 & 2490 & 104743 & 5749 & 3668 & 635 & 67349 & 392702 & 363871 & -\\
Dev & 1043&  277 & 5463 & 1500 & 408 & 71 & 872 & 9815/9832 & 40432 & -\\
Test & 1063 & 3000 & 5463 & 1379 & 1725 & 146 & 1821 & 9796/9847 & 390965 & 1104\\
    \bottomrule
    \end{tabular}
    \caption{GLUE dataset statistics.}
    \label{tab:data-stats}
\end{table*}

\subsection{Self-supervised Learning Tasks}
\label{Self-supervised Learning Tasks}

In this work, we use two self-supervised learning tasks -- masked token prediction (MTP) and sentence augmentation type prediction (SATP) -- to perform our studies while noting that other SSL tasks are also applicable. 

\begin{itemize}[leftmargin=*]
\item \textbf{Masked Token Prediction (MTP)}
This task is used in BERT. Some percentage of  input tokens are masked at random. Texts with masked tokens are fed into a text encoder which learns a latent representation for each token including the masked ones. The task is to predict these masked tokens by feeding  hidden vectors (produced by the encoder) corresponding to  masked tokens into an output softmax over word vocabulary.

\item \textbf{Sentence Augmentation Type Prediction (SATP)} Given an original text $o$, we apply different types of  augmentation methods to create augmented texts from $o$. We train a model to predict which type of augmentation was applied to an augmented text.  
We consider four types of augmentation operations used in~\citep{wei-zou-2019-eda}, including synonym replacement, random insertion, random swap, and random deletion. Synonym replacement randomly chooses  10\% of non-stop tokens from  original texts and replaces each of them with a randomly selected synonym. In random insertion, for a randomly chosen non-stop token in  a text, among the synonyms of this token, one randomly selected synonym is inserted into a random position in the text.  This operation is performed for 10\% of tokens.
Synonyms for synonym replacement and random insertion are obtained from Synsets in  NLTK~\citep{nltk} which are  constructed based on WordNet~\citep{Miller1995WordNetAL}. Synsets serve as a synonym dictionary containing  groupings of synonymous words. Some words have only one Synset and some have several. In synonym replacement, if a selected word in a sentence has multiple synonyms, we randomly choose one of them, and replace all occurrences of this word in the sentence with this synonym.
  Random swap randomly chooses two tokens in a text and swaps their positions. This operation is performed for 10\% of token pairs. Random deletion randomly removes a token with a probability of 0.1. 
  In this SSL task,  an augmented sentence is fed into a text encoder and the encoding is fed into a 4-way classification head to predict which operation was applied to generate this augmented sentence. 

\end{itemize}

\subsection{Text Encoder}

We use a Transformer encoder to perform the study while noting that other text encoders are also applicable. Different from sequence-to-sequence  models~\cite{sutskever2014sequence} that are based on recurrent neural networks (e.g., LSTM~\cite{hochreiter1997long}, GRU~\cite{chung2014empirical}) which model a sequence of tokens via a recurrent manner and hence is computationally inefficient, Transformer eschews recurrent computation and instead uses self-attention which not only can capture dependency between tokens but also is amenable for parallel computation with high efficiency. Self-attention calculates the correlation among every pair of tokens and uses these correlation scores to create ``attentive" representations by taking weighted summation of tokens' embeddings. Transformer is composed of building blocks, each consisting of a self-attention layer and a position-wise feed-forward layer.  Residual connection \cite{he2016deep} is applied around each of these two sub-layers, followed by layer normalization~\cite{ba2016layer}. Given an input sequence, an encoder -- which is a stack of such building blocks -- is applied to obtain a representation for each token.

\section{Experiments}
\subsection{Datasets}
We evaluated our method on the  datasets used in \citep{dontstoppretraining2020}, which are from  various domains. For each dataset, we follow the train/development/test split specified in  \citep{dontstoppretraining2020}. Dataset statistics are summarized in  Table \ref{tab:data_tasks}. 

In addition, we performed experiments on the datasets in the GLUE benchmark~\cite{wang2018glue}. The General Language Understanding Evaluation (GLUE) benchmark has 10 tasks, including 2 single-sentence tasks, 3 similarity and paraphrase tasks, and 5 inference tasks.  For each GLUE task, labels in  development sets are publicly available while those in test sets are not released. We obtain performance on  test sets by submitting inference results to  GLUE evaluation server\footnote{https://gluebenchmark.com/leaderboard}. Table~\ref{tab:data-stats} shows the statistics of data split in each task.

\subsection{Experimental Setup}
\subsubsection{Baselines} For experiments on datasets used in~\cite{dontstoppretraining2020},  text encoders in all methods are initialized using  pretrained RoBERTa~\cite{liu2019roberta}. For experiments on GLUE datasets,  text encoders  are initialized using  pretrained BERT~\cite{liu2019roberta} or pretrained RoBERTa.  
We compare our proposed SSL-Reg  with the following  baselines.
\begin{itemize}[leftmargin=*]
\item \textbf{Unregularized RoBERTa}~\cite{Liu2019RoBERTaAR}. 
In this approach, the Transformer encoder is initialized with pretrained RoBERTa. Then the pretrained encoder and a classification head form a text classification model, which is then finetuned on a target  classification task. Architecture of the classification model is the same as that in ~\cite{Liu2019RoBERTaAR}. Specifically,  representation of the \texttt{[CLS]} special token is passed to a feedforward layer for class prediction. Nonlinear activation function in the feedforward layer is tanh. During finetuning, no SSL-based regularization is used. This approach is evaluated on all datasets used in~\cite{dontstoppretraining2020} and all datasets in GLUE.

\item  \textbf{Unregularized BERT.}  
This approach is the same as  unregularized RoBERTa, except that the Transformer encoder is initialized by  pretrained BERT~\cite{devlin2018bert} instead of RoBERTa. This approach is evaluated on all GLUE datasets. 

    \item  \textbf{Task adaptive pretraining (TAPT)}~\cite{dontstoppretraining2020}.
In this approach, given the   Transformer encoder pretrained using RoBERTa or BERT on large-scale external corpora, it is further pretrained by RoBERTa or BERT on  input texts in a target classification dataset (without using class labels). Then this further pretrained encoder is used to initialize the encoder in the text classification model and is  finetuned to perform  classification tasks which use both  input texts and their class labels. Similar to SSL-Reg, TAPT also performs SSL on  texts in  target classification dataset. The difference is: TAPT performs  SSL task and  classification task sequentially while SSL-Reg performs these two tasks jointly. TAPT is studied for all  datasets in this paper.

    \item  \textbf{Domain adaptive pretraining 
    (DAPT)}~\cite{dontstoppretraining2020}.
In this approach,  given a pretrained encoder on  large-scale external corpora, the encoder is further pretrained on a small-scale corpora whose domain is similar to that of texts in a target classification dataset. Then this further pretrained encoder is finetuned in a  classification task. DAPT is similar to TAPT, except that TAPT performs the second stage pretraining on texts $T$ in the classification dataset while  DAPT performs the second stage pretraining on external texts whose domain is similar to that of $T$ rather than directly on $T$. The external dataset is usually much larger than $T$.

 \item  \textbf{TAPT+SSL-Reg.} When finetuning the classification model, SSL-Reg is applied. The rest is the same as TAPT. 
  \item  \textbf{DAPT+SSL-Reg.} When finetuning the classification model, SSL-Reg is applied. The rest is the same as DAPT.
    
\end{itemize}

\begin{table}[t]
\small
\begin{center}
\resizebox{0.45\textwidth}{!}
{
\begin{tabular}{c|ccc}
\toprule
\multirow{2}{*}{Task} 
&  Epoch & \makecell{Learning \\ Rate} & \makecell{Regularization \\ Parameter} \\
\hline
CoLA    & 10    & 3e-5      & 0.2    \\  
SST-2    & 3    & 3e-5      & 0.05  \\  
MRPC    & 5    & 4e-5      & 0.05   \\  
STS-B    & 10    & 4e-5      & 0.1  \\  
QQP    & 5    & 3e-5      & 0.2 \\  
MNLI    & 3    & 3e-5      & 0.1    \\  
QNLI    & 4    & 4e-5      & 0.5    \\  
RTE    & 10    & 3e-5      & 0.1 \\  
WNLI    & 5    & 5e-5      & 2\\ 
\bottomrule
\end{tabular}
} 
\end{center}
\caption{Hyperparameter settings for BERT on GLUE datasets, where the SSL task is MTP.}
\label{tab:GLUE_finetune_settings_MTP}
\end{table}

\begin{table}[t]
\small
\begin{center}
\resizebox{0.45\textwidth}{!}
{
\begin{tabular}{c|ccc}
\toprule
\multirow{2}{*}{Task}   
&  Epoch & \makecell{Learning \\ Rate} & \makecell{Regularization \\ Parameter} \\
\hline
CoLA    & 6    & 3e-5   &0.4    \\  
SST-2    & 3    & 3e-5  &0.8    \\  
MRPC    & 5    & 4e-5     &0.05 \\  
STS-B    & 10    & 4e-5    &0.05   \\  
QQP    & 5    & 3e-5   & 0.4 \\  
MNLI    & 4    & 3e-5   & 0.5  \\  
QNLI    & 4    & 4e-5    & 0.05  \\  
RTE    & 8    & 3e-5    & 0.6   \\  
WNLI    & 5    & 5e-5   & 0.1   \\ 
\bottomrule
\end{tabular}
}
\end{center}
\caption{Hyperparameter settings for BERT on GLUE datasets, where the SSL task is SATP.}
\label{tab:GLUE_finetune_settings_SATP}
\end{table}

\begin{table}[t]
\small
\begin{center}
\resizebox{0.45\textwidth}{!}
{
\begin{tabular}{c|ccc}
\toprule
\multirow{2}{*}{Task} 
&  Epoch & \makecell{Learning \\ Rate} & \makecell{Regularization \\ Parameter} \\
\hline
CoLA    & 10    & 1e-5      & 0.8    \\  
SST-2    & 3    & 1e-5      & 1.0  \\  
MRPC    & 10    & 1e-5      & 0.01   \\  
STS-B    & 10    & 2e-5      & 0.01  \\  
QQP    & 10    & 1e-5      & 0.1 \\  
MNLI    & 3    & 1e-5      & 0.1    \\  
QNLI    & 3    & 1e-5      & 0.1    \\  
RTE    & 10    & 2e-5      & 0.1 \\  
WNLI    & 10    & 2e-5      & 0.02\\ 
\bottomrule
\end{tabular}
}
\end{center}
\caption{Hyperparameter settings for RoBERTa on GLUE datasets, where the SSL task is MTP.}
\label{tab:RoBERTa_GLUE_finetune_settings_MTP}
\end{table}

\begin{table*}[t]
\small
\centering
\begin{tabular}{llcccccccc}
\toprule
\textbf{Dataset} 
& \textbf{RoBERTa} & \textbf{DAPT} & \textbf{TAPT} & \textbf{SSL-Reg} & \textbf{TAPT+SSL-Reg} & \textbf{DAPT+SSL-Reg} \\ 

\midrule 
\chemprot 
& 81.9$_{1.0}$& 84.2$_{0.2}$& 82.6$_{0.4}$&83.1$_{0.5}$& 83.5$_{0.1}$& \bf{84.4}$_{0.3}$ & \\ 
\rct  
& 87.2$_{0.1}$ & 87.6$_{0.1}$ & \bf 87.7$_{0.1}$ & 87.4$_{0.1}$& \bf{87.7}$_{0.1}$& \bf{87.7}$_{0.1}$ \\  
\midrule[0.03em]
\arccite 
& 63.0$_{5.8}$& 75.4$_{2.5}$ & 67.4$_{1.8}$& 69.3$_{4.9}$ & 68.1$_{2.0}$ & \bf{75.7}$_{1.4}$ \\ 
\sciie
& 77.3$_{1.9}$& 80.8$_{1.5}$& 79.3$_{1.5}$ & 81.4$_{0.8}$ & 80.4$_{0.6}$ & \bf{82.3}$_{0.8}$ \\ 
\midrule[0.03em]
\hp 
& 86.6$_{0.9}$& 88.2$_{5.9}$& 90.4$_{5.2}$ & 92.3$_{1.4}$ & \bf{93.2}$_{1.8}$ & 90.7$_{3.2}$ \\ 
\ag 
& 93.9$_{0.2}$& 93.9$_{0.2}$& \bf 94.5$_{0.1}$& 94.2$_{0.1}$ & 94.4$_{0.1}$& 94.0$_{0.1}$ \\
\midrule[0.03em]
\helpful
& 65.1$_{3.4}$& 66.5$_{1.4}$& 68.5$_{1.9}$& 69.4$_{0.2}$ &\bf{71.0}$_{1.0}$&68.3$_{1.4}$ \\
\imdb 
& 95.0$_{0.2}$& 95.4$_{0.1}$ & 95.5$_{0.1}$ & 95.7$_{0.1}$ & \bf{96.1}$_{0.1}$ & 95.4$_{0.1}$ \\
\bottomrule
\end{tabular}
\caption{Results on datasets used in \citep{dontstoppretraining2020}. For  vanilla (unregularized) RoBERTa, DAPT, and TAPT,  results are taken from \citep{dontstoppretraining2020}. For each method on each dataset, we  run it for four times with different random seeds.  Results are in  $m_s$ format, where $m$ denotes  mean and $s$ denotes  standard derivation. 
Following \citep{dontstoppretraining2020}, for \chemprot and \rct, we report micro-F1; for other datasets, we report macro-F1. 
}
\label{tab:main}
\end{table*}
\begin{table}[t]
\centering
\begin{tabular}{lcc}
\toprule
\textbf{Dataset} & \textbf{RoBERTa} & \textbf{SSL-Reg}   \\
\midrule
\chemprot & \bf 13.05 & 13.57  \\
\arccite & 28.67 & \bf 25.24  \\
\sciie & 19.51 & \bf 18.23  \\
\hp & 7.44 & \bf 5.64 \\
\bottomrule
\end{tabular}
\caption{Difference between F1 score on  training set and F1 score on  test set with or without SSL-Reg (MTP). \textbf{Bold} denotes a smaller difference, which means overfitting is less severe.}
\label{tab:difference}
\end{table}

\subsubsection{Hyperparameter Settings}
Hyperparameters were tuned on  development datasets.
\vspace{-0.3cm}
\paragraph{Hyperparameter settings for RoBERTa on datasets used in} \citep{dontstoppretraining2020}. 
For a fair comparison, most of our hyperparameters are the same as those in \citep{dontstoppretraining2020}. The maximum text length was set to 512.
 Text encoders in all methods are initialized using pretrained RoBERTa~\cite{liu2019roberta} on a large-scale external dataset. For TAPT, DAPT, TAPT+SSL-Reg, and DAPT+SSL-Reg, the second-stage pretraining on texts $T$ in a target classification dataset or on external texts whose domain is similar to that of $T$ is based on the pretraining approach in RoBERTa. 
    In SSL-Reg, the SSL task is masked token prediction. SSL loss function only considers the prediction of  masked tokens and ignores the prediction of  non-masked tokens.  Probability for  masking tokens is 0.15. 
If a token $t$ is chosen to be masked, 80\% of the time, we replace $t$  with a special token \texttt{[MASK]}; 10\% of the time, we replace $t$ with a  random word; and for the rest 10\% of the time, we keep $t$ unchanged.
For the regularization parameter in SSL-Reg, we set it to 0.2 for \arccite, 0.1 for \sciie, \chemprot, \ag, \rct, \helpful, \imdb, and 0.01 for \hp. 
For \arccite, \chemprot, \rct, \sciie and \hp, we trained SSL-Reg for 10 epochs; for \helpful, 5 epochs; for \ag, \rct and \imdb,  3 epochs. For all  datasets, we used a batch size of 16 with gradient accumulation.  We used  the AdamW optimizer \citep{Loshchilov2017FixingWD} with a warm-up proportion of 0.06, a weight decay of 0.1, and an epsilon of 1e-6. In AdamW,  $\beta_1$ and $\beta_2$ are set to 0.9 and 0.98, respectively. The maximum learning rate was 2e-5. 

\vspace{-0.3cm}
\paragraph{Hyperparameter settings for BERT on GLUE datasets.}
The maximum text length was set to 128. Since  external texts whose domains are similar to those of the GLUE texts are not available, we did not compare with DAPT and DAPT+SSL-Reg. For each method applied, text encoder is initialized using pretrained BERT~\cite{devlin2018bert} (with 24 layers) on a large-scale external dataset. In TAPT, the second-stage pretraining is performed using BERT. As we will show later on, TAPT does not perform well on GLUE datasets; therefore, we did not apply  TAPT+SSL-Reg on these datasets further.  
In SSL-Reg, we studied two SSL tasks: masked token prediction (MTP) and sentence augmentation type prediction (SATP). 
In MTP, we randomly mask 15\% of tokens in each text. 
 Batch size was set to 32 with gradient accumulation. We use the AdamW optimizer \citep{Loshchilov2017FixingWD} with a warm-up proportion of 0.1, a weight decay of 0.01, and an epsilon of 1e-8. In AdamW, $\beta_1$ and $\beta_2$ are set to 0.9 and 0.999, respectively. Other hyperparameter settings are presented in Table~\ref{tab:GLUE_finetune_settings_MTP} and Table~\ref{tab:GLUE_finetune_settings_SATP}.

\paragraph{Hyperparameter settings for RoBERTa on GLUE datasets.}
Most hyperparameter settings follow those in RoBERTa experiments performed  on datasets used in~\citep{dontstoppretraining2020}. 
 
We set different learning rates and different epoch numbers for different datasets as guided by \cite{Liu2019RoBERTaAR}. In addition, we set different regularization parameters for different datasets. These hyperparameters are listed in Table~\ref{tab:RoBERTa_GLUE_finetune_settings_MTP}.

\begin{figure*}[t]
    \centering
    \vspace{-0.2cm}
    \includegraphics[width=1.0\textwidth]{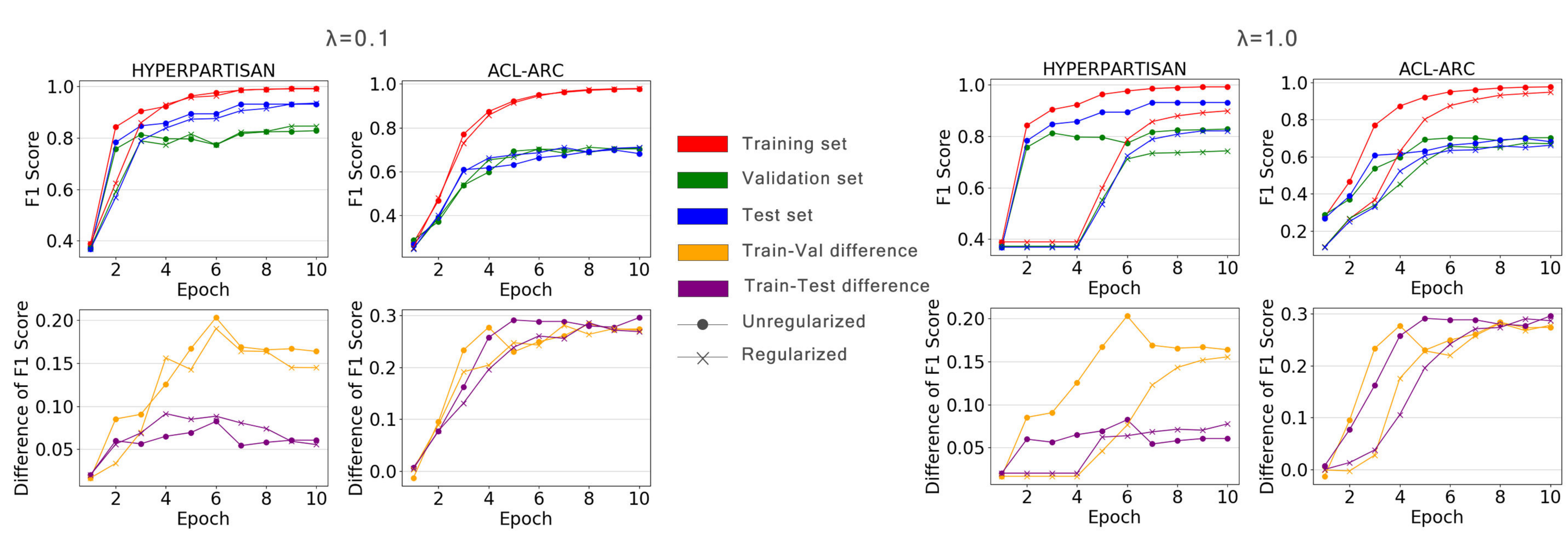}
    \vspace{-0.2cm}
    \caption{Training dynamics of unregularized RoBERTa and SSL-Reg (denoted by ``Regularized'') on \hp and \arccite. In SSL-Reg, we experimented with two values of the regularization parameter $\lambda$: 0.1 and 1. 
    }
    \label{fig:dynamics}
\end{figure*}

\subsection{Results}

\subsubsection{Results on the datasets used in~\cite{dontstoppretraining2020}.}
Performance of text classification on  datasets used in~\cite{dontstoppretraining2020} is reported in Table \ref{tab:main}. Following \citep{dontstoppretraining2020}, for \chemprot and \rct, we report micro-F1; for other datasets, we report macro-F1. 
From this table, we make the following observations. \textbf{First}, SSL-Reg outperforms unregularized RoBERTa significantly on all datasets. We used a double-sided t-test to perform significance tests. The p-values are less than 0.01, which indicate strong statistical significance. 
This demonstrates the effectiveness of our proposed SSL-Reg approach in alleviating overfitting and improving generalization performance. To further confirm this, we measure the difference between F1 scores on the training set and test set in Table \ref{tab:difference}. A larger difference implies more overfitting: performing well on the training set and less well on the test set. As can  be seen, the train-test difference under SSL-Reg is smaller than that under RoBERTa. SSL-Reg encourages  text encoders to solve an additional task based on SSL, which reduces the risk of overfitting to the data-deficient classification task on  small-sized training data. In Figure~\ref{fig:dynamics}, we compare the training dynamics of unregularized RoBERTa and SSL-Reg (denoted by ``Regularized"). As can be seen, under a large regularization parameter $\lambda=1$, our method achieves smaller differences between training accuracy and validation accuracy than unregularized RoBERTa; our method also achieves smaller differences between training accuracy and test accuracy than unregularized RoBERTa. These results show that our proposed SSL-Reg indeed acts as a regularizer which reduces the gap between performances on training set and validation/test set. Besides, when increasing $\lambda$ from 0.1 to 1, the training accuracy of SSL-Reg decreases considerably. This also indicates that SSL-Reg acts as a regularizer which penalizes training performance. 
\textbf{Second}, on 6 out of the 8 datasets, SSL-Reg performs better than TAPT. On the other two datasets, SSL-Reg is on par with TAPT. This shows that SSL-Reg is more effective than TAPT. SSL-Reg and TAPT both leverage  input texts in  classification datasets for self-supervised learning. The difference is: TAPT uses these texts to pretrain the  encoder while SSL-Reg uses these texts to regularize the encoder during finetuning. In SSL-Reg, the encoder is learned to perform  classification tasks and SSL tasks simultaneously. Thus the encoder is not completely biased to  classification tasks. In TAPT, the encoder is first learned by performing  SSL tasks, then finetuned by performing  classification tasks. There is a risk that after finetuning, the encoder is largely biased to  classification tasks on  small-sized training data, which leads to overfitting. \textbf{Third}, on 5 out of the 8 datasets, SSL-Reg performs better than DAPT, although DAPT leverages additional external data. The reasons are two-fold: 1) similar to TAPT, DAPT performs  SSL task first and then  classification task separately; as a result,  the encoder may be eventually biased to  classification task on  small-sized training data; 2)  external data used in DAPT still has a domain shift with  target dataset; this domain shift may render the text encoder pretrained on  external data not suitable for  target task. To verify this, we measure the domain similarity between external texts and target texts by calculating cosine similarity between the BERT embeddings of these texts. The similarity score is 0.14. As a reference, the similarity score between texts in the target dataset is 0.27. This shows that there is indeed a domain difference between external texts and target texts. \textbf{Fourth}, on 6 out of  8 datasets, TAPT+SSL-Reg performs better than TAPT. On the other two datasets, TAPT+SSL-Reg is on par with TAPT. This further demonstrates the effectiveness of SSL-Reg.  \textbf{Fifth}, on all eight  datasets, DAPT+SSL-Reg performs better than DAPT.  This again shows that SSL-Reg is effective.  \textbf{Sixth}, on 6 out of  8 datasets, TAPT+SSL-Reg performs better than SSL-Reg, indicating that it is beneficial to use both TAPT and SSL-Reg: first use the target texts to pretrain the encoder based on SSL, then apply  SSL-based regularizer on these target texts during finetuning. \textbf{Seventh}, DAPT+SSL-Reg performs better than SSL-Reg on 4 datasets, but worse on the other 4 datasets, indicating that with SSL-Reg used, DAPT is not necessarily useful. 
\textbf{Eighth},  on smaller datasets, improvement achieved by SSL-Reg over baselines is larger. For example, on \hp which has only about 500 training examples, improvement of SSL-Reg over RoBERTa is 5.7\% (absolute percentage). Relative improvement is 6.6\%. As another example, on \arccite which has about 1700 training examples, improvement of SSL-Reg over RoBERTa is 6.3\% (absolute percentage). Relative improvement is 10\%. In contrast, on large datasets such as \rct which contains about 180000 training examples, improvement of  SSL-Reg over RoBERTa is 0.2\% (absolute percentage). Relative improvement is 0.2\%. On another large dataset \ag which contains 115000 training examples,  improvement of  SSL-Reg over RoBERTa is 0.3\% (absolute percentage). Relative improvement is 0.3\%. The reason that SSL-Reg achieves better improvement on smaller datasets is that smaller datasets are more likely to lead to overfitting and SSL-Reg is more needed to alleviate this overfitting.

\begin{figure}[t]
    \centering
    \includegraphics[width=0.52\textwidth]{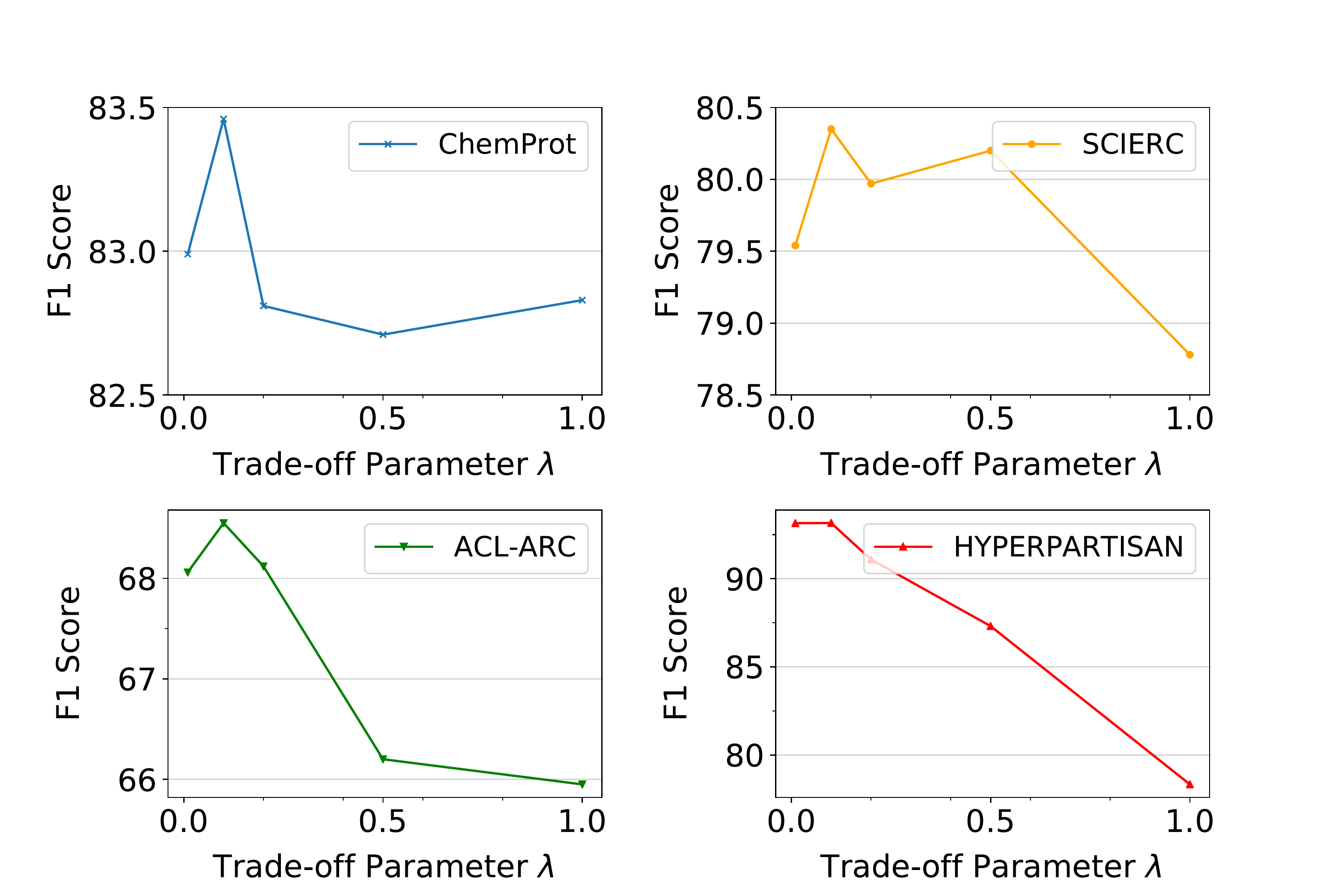}
    \caption{How regularization parameter in SSL-Reg affects text classification F1 score.}
    \label{fig:joint_weight}
\end{figure}

\begin{table*}[t]
    \centering
    \small
    \begin{tabular}{lccccc}
        \toprule
        & \textbf{\makecell{CoLA \\ (Matthew Corr.)}}    & \textbf{\makecell{SST-2 \\ (Accuracy)}}   & \textbf{\makecell{RTE \\ (Accuracy)}}   & \textbf{\makecell{QNLI \\ (Accuracy)}} & \textbf{\makecell{MRPC \\ (Accuracy/F1)}} \\  \midrule 
        \emph{The median result} \\
        BERT, Lan et al. 2019   & 60.6  & 93.2  & 70.4  & 92.3   & \textbf{88.0/-} \\
        BERT, our run  & 62.1   & 93.1 & 74.0   & 92.1 & 86.8/90.8   \\ 
        TAPT & 61.2 & 93.1  & 74.0  & 92.0  & 85.3/89.8 \\
        SSL-Reg (SATP) &63.7   & \textbf{93.9}   &\textbf{74.7}   & 92.3  &86.5/90.3  \\
        SSL-Reg (MTP) & \textbf{63.8} & 93.8 & \textbf{74.7} & \textbf{92.6} & 87.3/90.9 \\ 
        \midrule
        \emph{The best result} \\
        BERT, our run   & 63.9  & 93.3  & 75.8  & 92.5  & \textbf{89.5/92.6} \\
        TAPT & 62.0 & 93.9  & 76.2  & 92.4  & 86.5/90.7 \\
        SSL-Reg (SATP) &65.3   &94.6   &\textbf{78.0}   & 92.8  &88.5/91.9  \\
        SSL-Reg (MTP) & \textbf{66.3} & \textbf{94.7} & \textbf{78.0} & \textbf{93.1} & \textbf{89.5}/92.4 \\
        \bottomrule
    \end{tabular}
\caption{Results of BERT-based experiments on  GLUE development sets, where  results on MNLI and QQP are the median of five runs and  results on other datasets are the median of nine runs. The size of MNLI and QQP is very large, taking a long time to train on. Therefore, we reduced the number of runs. Because we used a different optimization method to re-implement BERT, our median performance is not the same as that reported in \cite{lan2019albert}.
}
\label{tab:glue_dev_first}
\end{table*}

\begin{table*}[t]
    \centering
    \small
    \begin{tabular}{lcccc}
        \toprule
        & \textbf{\makecell{MNLI-m/mm \\ (Accuracy)}}    & \textbf{\makecell{QQP \\ (Accuracy/F1)}}   & \textbf{\makecell{STS-B \\ (Pearson Corr./Spearman Corr.) }}   & \textbf{\makecell{WNLI \\ (Accuracy)}}  \\  \midrule
        \emph{The median result} \\
        BERT, Lan et al. 2019   & 86.6/-  & 91.3/-   & 90.0/-  & -  \\
        BERT, our run  & 86.2/86.0   & 91.3/88.3 & 90.4/90.0   & 56.3  \\ 
        TAPT & 85.6/85.5    & 91.5/88.7 & 90.6/90.2 & 53.5\\
        SSL-Reg (SATP) & 86.2/86.2   & 91.6/88.8   &\textbf{90.7/90.4}    & 56.3  \\
        SSL-Reg (MTP) & \textbf{86.6/86.6} & \textbf{91.8/89.0} & 90.7/90.3 & 56.3  \\
        \midrule
        \emph{The best result} \\
        BERT, our run   & 86.4/86.3  & 91.4/88.4  & 90.9/90.5  & 56.3 \\
        TAPT    & 85.7/85.7 & 91.7/89.0 & 90.8/90.4 & 56.3\\
        SSL-Reg (SATP) & 86.4/86.5   & 91.8/88.9   &\textbf{91.1/90.8}    & \textbf{59.2}  \\
        SSL-Reg (MTP) & \textbf{86.9/86.9} & \textbf{91.9/89.1}  & \textbf{91.1/90.8} & 57.7 \\
        \bottomrule
    \end{tabular}
\caption{
Continuation of Table~\ref{tab:glue_dev_first}.
}
\label{tab:glue_dev_second}
\end{table*}

\begin{table*}[t]
    \centering
    \begin{tabular}{lcccc}
        \toprule
        & \textbf{BERT}  & \textbf{TAPT}  & \textbf{SSL-Reg (SATP)}  & \textbf{SSL-Reg (MTP)}\\ \midrule
        CoLA (Matthew Corr.) & 60.5 & 61.3 & \textbf{63.0}  & 61.2\\
        SST-2 (Accuracy) & 94.9 & 94.4 & 95.1  & \textbf{95.2}\\
        RTE (Accuracy)   & 70.1  & 70.3   & 71.2  & \textbf{72.7}\\
        QNLI (Accuracy)   & 92.7  & 92.4   & 92.5  & \textbf{93.2}\\
        MRPC (Accuracy/F1)  & 85.4/89.3   & 85.9/89.5    & 85.3/89.3  & \textbf{86.1/89.8} \\
        MNLI-m/mm (Accuracy) & \textbf{86.7}/85.9  & 85.7/84.4   & 86.2/85.4  & 86.6/\textbf{86.1} \\
        QQP (Accuracy/F1)    & 89.3/72.1  & 89.3/71.6   & 89.6/72.2  & \textbf{89.7/72.5}\\
        STS-B (Pearson Corr./Spearman Corr.) & 87.6/86.5  & \textbf{88.4}/87.3   & 88.3/\textbf{87.5}  & 88.1/87.2     \\
        WNLI (Accuracy)  & 65.1   & 65.8  & 65.8  & \textbf{66.4}\\
        AX(Matthew Corr.)  & 39.6  & 39.3   & 40.2  & \textbf{40.3}\\ \midrule
         \textbf{Average} & 80.5    & 80.6  & 81.0  & \textbf{81.3}\\
        \bottomrule
    \end{tabular}
\caption{Results of BERT-based experiments on GLUE test sets, which are scored by the GLUE evaluation server (\url{https://gluebenchmark.com/leaderboard}). 
Models evaluated on AX are trained on the  training dataset of MNLI.
}
\label{tab:glue_test}
\end{table*}

\begin{table*}[t]
    \centering
    \small
    \begin{tabular}{lccccc}
        \toprule
        & \textbf{\makecell{CoLA \\ (Matthew Corr.)}}    & \textbf{\makecell{SST-2 \\ (Accuracy)}}   & \textbf{\makecell{RTE \\ (Accuracy)}}   & \textbf{\makecell{QNLI \\ (Accuracy)}} & \textbf{\makecell{MRPC \\ (Accuracy/F1)}} \\  \midrule 
        \emph{The median result} \\
        RoBERTa, Liu et al. 2019   & 68.0  & \textbf{96.4}  & \textbf{86.6}   & 94.7   & \textbf{90.9/-} \\
        RoBERTa, our run  & 68.7   & 96.1 & 84.8   & 94.6 & 89.5/92.3   \\ 
        SSL-Reg (MTP) & \textbf{69.2} & 96.3 & 85.2 & \textbf{94.9} & 90.0/\textbf{92.7} \\ 
        \midrule
        \emph{The best result} \\
        RoBERTa, our run   & 69.2  & 96.7  & 86.6 & 94.7  & 90.4/93.1 \\
        SSL-Reg (MTP) & \textbf{70.2} & 96.7 & 86.6 & \textbf{95.2} & \textbf{91.4}/\textbf{93.8} \\
        \bottomrule
    \end{tabular}
\caption{Results of RoBERTa-based experiments on GLUE development sets, where the median results are the median of five runs. Because we used a different optimization method to re-implement RoBERTa, our median performance is not the same as that reported in \cite{Liu2019RoBERTaAR}. 
}
\label{tab:roberta_glue_dev_first}
\end{table*}

\begin{table*}[t]
    \centering
    \small
    \begin{tabular}{lcccc}
        \toprule
        & \textbf{\makecell{MNLI-m/mm \\ (Accuracy)}}    & \textbf{\makecell{QQP \\ (Accuracy)}}   & \textbf{\makecell{STS-B \\ (Pearson Corr./Spearman Corr.) }}   & \textbf{\makecell{WNLI \\ (Accuracy)}}  \\  \midrule
        \emph{The median result} \\
        RoBERTa, Liu et al. 2019   & 90.2/90.2  & 92.2   & 92.4/-  & -  \\
        RoBERTa, our run  & 90.5/90.5   & 91.6 & 92.0/92.0   & 56.3  \\ 
        SSL-Reg (MTP) & \textbf{90.7/90.7} & 91.6 & 92.0/92.0 & \textbf{62.0}  \\
        \midrule
        \emph{The best result} \\
        RoBERTa, our run   & 90.7/90.5  & 91.7  & 92.3/92.2  & 60.6 \\
        SSL-Reg (MTP) & 90.7/90.5 & \textbf{91.8}  & 92.3/92.2 & \textbf{66.2} \\
        \bottomrule
    \end{tabular}
\caption{
Continuation of Table~\ref{tab:roberta_glue_dev_first}.
}
\label{tab:roberta_glue_dev_second}
\end{table*}

\begin{table*}[t]
    \centering
    \small
    \begin{tabular}{lcccccc}
        \toprule
        & \textbf{\makecell{CoLA}}    & \textbf{\makecell{SST-2 }}   & \textbf{\makecell{RTE}}   & \textbf{\makecell{QNLI}} & \textbf{\makecell{MRPC}}& \textbf{\makecell{STS-B}} \\  \midrule 
        SR+RD+RI+RS   & \textbf{63.6}  & \textbf{94.0}  & \textbf{74.8}   & \textbf{92.2}   & 86.8/90.6 & \textbf{90.6/90.3} \\
        SR+RD+RI  & 63.4   & 93.8 & 72.8   & 92.1 & 86.9/90.8 & 90.6/90.2   \\ 
        SR+RD & 61.6 & 93.6 & 72.5 & \textbf{92.2} &  \textbf{87.2}/\textbf{91.0} & \textbf{90.6}/\textbf{90.3} \\ 
        \bottomrule
    \end{tabular}
\caption{Ablation study on sentence augmentation types in SSL-Reg (SATP), where SR, RD, RI and RS denotes synonym replacement, random deletion, random insertion, and random swap respectively. Results are averaged over 5  runs with different random initialization.}
\label{tab:ablation_SATP}
\end{table*}

Figure~\ref{fig:joint_weight} shows how  classification F1 score varies as we increase regularization parameter $\lambda$ from 0.01 to 1.0 in SSL-Reg.
As can be seen, starting from 0.01, when the regularizer parameter is increasing,  F1 score increases. This is because a larger $\lambda$ imposes a stronger regularization effect, which helps to reduce overfitting. However, if $\lambda$ becomes too large,  F1 score drops. This is because the regularization effect is too strong, which dominates  classification loss. Among these 4 datasets,  F1 score drops dramatically on \hp as $\lambda$ increases. This is probably because this dataset contains very long sequences. This makes MTP on this dataset more difficult and therefore yields an excessively strong regularization outcome that hurts classification performance. Compared with \hp, F1 score is less sensitive on other datasets because their sequence lengths are relatively smaller.

\subsubsection{Results on the GLUE benchmark}

Table~\ref{tab:glue_dev_first} and Table~\ref{tab:glue_dev_second} show results of BERT-based experiments on development sets of GLUE. 
As mentioned in~\citep{devlin-etal-2019-bert}, for the 24-layer version of BERT, finetuning is sometimes unstable on small datasets, so we run each method several times and report the median and best performance. Table~\ref{tab:glue_test} shows the best performance on  test sets. 
  Following~\cite{wang2018glue}, we report  Matthew  correlation on CoLA, Pearson correlation and Spearman correlation on STS-B, accuracy and F1 on MRPC and QQP. For the rest datasets, we report accuracy. From these tables, we make the following observations. \textbf{First}, SSL-Reg methods including SSL-Reg-SATP and SSL-Reg-MTP outperform unregularized BERT (our run) on most datasets: 1) on  test sets, SSL-Reg-SATP performs better than BERT on 7 out of  10 datasets and SSL-Reg-MTP performs better than BERT on  9 out of 10 datasets; 2) in terms of median results on  development sets, SSL-Reg-SATP performs better than BERT (our run) on 7 out of  9 datasets and SSL-Reg-MTP performs better than BERT (our run) on 8 out of  9 datasets; 3) in terms of best results on  development sets, SSL-Reg-SATP performs better than BERT (our run) on 8 out of  9 datasets and SSL-Reg-MTP performs better than BERT (our run) on 8 out of  9 datasets. This further demonstrates the effectiveness of SSL-Reg in improving generalization performance. \textbf{Second}, on 7 out of  10 test sets, SSL-Reg-SATP outperforms TAPT; on  8 out of  10 test sets, SSL-Reg-MTP outperforms TAPT. On most development datasets, SSL-Reg-SATP and SSL-Reg-MTP outperform TAPT. The only exception is: on  QQP development set, the best F1 of TAPT is slightly better than that of SSL-Reg-SATP.  This further demonstrates that performing SSL-based regularization on target texts is more effective than using them for pretraining. \textbf{Third}, overall, SSL-Reg-MTP performs better than SSL-Reg-SATP. For example, on 8 out of  10 test datasets, SSL-Reg-MTP performs better than SSL-Reg-SATP. MTP works better than SATP probably because it is a more challenging self-supervised learning task that encourages encoders to learn more powerful representations. \textbf{Fourth},  improvement of SSL-Reg methods over BERT is  more prominent on smaller training datasets, such as CoLA and RTE. This may be because smaller training datasets are more likely to lead to overfitting where the advantage of  SSL-Reg in alleviating overfitting can be better played. 

Table~\ref{tab:roberta_glue_dev_first} and~\ref{tab:roberta_glue_dev_second}  show results of RoBERTa-based experiments on development sets of GLUE. 
From these two tables, we make observations that are similar to those in Table~\ref{tab:glue_dev_first} and Table~\ref{tab:glue_dev_second}. In terms of median results, SSL-Reg (MTP) performs better than unregularized RoBERTa (our run) on 7 out of 9 datasets and achieves the same performance as RoBERTa (our run) on the rest 2 datasets. In terms of best results, SSL-Reg (MTP) performs better than RoBERTa (our run) on 5 out of  9 datasets and achieves the same performance as RoBERTa (our run) on the rest 4 datasets. This further demonstrates the effectiveness of our proposed SSL-Reg approach which uses an  MTP-based self-supervised task to regularize the finetuning of RoBERTa.

In SSL-Reg (SATP), we perform an ablation study on different types of sentence augmentation. Results are shown in Table~\ref{tab:ablation_SATP}, where SR, RD, RI and RS denote synonym replacement, random deletion, random insertion, and random swap, respectively. SR+RD+RI+RS means that we apply these four types of operations to augment sentences; given an augmented sentence $a$, we predict which of the four types of operations was applied to an original sentence to create $a$. SR+RD+RI+RS and SR+RD hold similar meanings. From this table, we make the following observations. \textbf{First}, as the number of augmentation types increases from 2 (SR+RD) to 3 (SR+RD+RI) then to 4 (SR+RD+RI+RS), the performance increases in general. This shows that it is beneficial to have more augmentation types in SATP. The reason is that more types make the SATP task more challenging and solving a more challenging self-supervised learning task can enforce  sentence encoders to learn more powerful representations. \textbf{Second}, SR+RD+RI+RS outperforms SR+RD+RI on 5 out of  6 datasets. This demonstrates that leveraging random swap (RS) for SATP can learn more effective representations of sentences. The reason is: SR, RD, and RI change the collection of  tokens in a sentence via  synonym replacement, random deletion, and random insertion; RS does not change the collection of tokens, but changes the order of tokens; therefore, RS is complementary to the other three operations; adding RS can bring in additional benefits that are complementary to those of SR, RD, and RI. \textbf{Third}, SR+RD+RI performs much better than SR+RD on CoLA and is on par with SR+RD on the rest five datasets. This shows that adding RI to SR+RD is beneficial. Unlike synonym replacement (SR) and random deletion (RD) which do not increase the number of tokens in a sentence, RI increases  token number. Therefore, RI is complementary to SR and RD and can bring in additional benefits.

\section{Conclusions and Future Work}
In this paper, we propose to use self-supervised learning to alleviate overfitting in text classification problems. We propose SSL-Reg, which is a regularizer based on SSL and a text encoder is trained to simultaneously minimize classification loss and  regularization loss. We demonstrate the effectiveness of our methods on 17 text classification datasets.

For future works, we will use other self-supervised learning tasks to perform regularization, such as contrastive learning, which predicts whether two augmented sentences stem from the same original sentence.

\bibliography{tacl2018}
\bibliographystyle{acl_natbib}

\end{document}